\documentclass{article}
\usepackage{spconf,amsmath,graphicx,hyperref} % 
\usepackage{amssymb}  % 
\usepackage{enumitem}

\usepackage{amsfonts}   
\usepackage{multirow}
\usepackage{tabularx}
\usepackage{booktabs}   % 
\usepackage{makecell}   % 
\usepackage{array}
\usepackage{ragged2e}

\usepackage{caption}
\title{A Multimodal Cross-View Model for Predicting Postoperative Neck Pain in Cervical Spondylosis Patients}
\name{%
	\begin{tabular}[t]{c}
		Jingyang Shan\textsuperscript{1,2,*} \quad
		Qishuai Yu\textsuperscript{3,4,*} \quad
		Jiacen Liu\textsuperscript{1,2} \quad
		Shaolin Zhang\textsuperscript{1,2} \quad
		Wen Shen\textsuperscript{1,2} \\
		Yanxiao Zhao\textsuperscript{1,2} \quad
		Tianyi Wang\textsuperscript{1,2} \quad
		Xin Lan\textsuperscript{1,2} \quad
		Xiaolin Qin\textsuperscript{1,2} \quad
		Yiheng Yin\textsuperscript{3}
	\end{tabular}%
  \thanks{* These authors contributed equally.}%
}

\address{$^{1}$ Chengdu Institute of Computer Applications, Chinese Academy of Sciences, Chengdu, China. \\
      $^{2}$ University of Chinese Academy of Sciences, Beijing, China.\\
	  $^{3}$ Department of Neurosurgery, the First Medical Center, Chinese PLA General Hospital, Beijing, China.\\
	  $^{4}$ School of Medicine, Nankai University, Tianjin, China.\\
	  }
\begin{document}
\maketitle

\begin{abstract}
Neck pain is the primary symptom of cervical spondylosis, yet its underlying mechanisms remain unclear, leading to uncertain treatment outcomes.  
To address the challenges of multimodal feature fusion caused by imaging differences and spatial mismatches, this paper proposes an Adaptive Bidirectional Pyramid Difference Convolution (ABPDC) module that facilitates multimodal integration by exploiting the advantages of difference convolution in texture extraction and grayscale invariance, and a Feature Pyramid Registration Auxiliary Network (FPRAN) to mitigate structural misalignment.  
Experiments on the MMCSD dataset demonstrate that the proposed model achieves superior prediction accuracy of postoperative neck pain recovery compared with existing methods, and ablation studies further confirm its effectiveness.

\end{abstract}

\begin{figure*}
 \centering
  \includegraphics[width=0.9\linewidth]{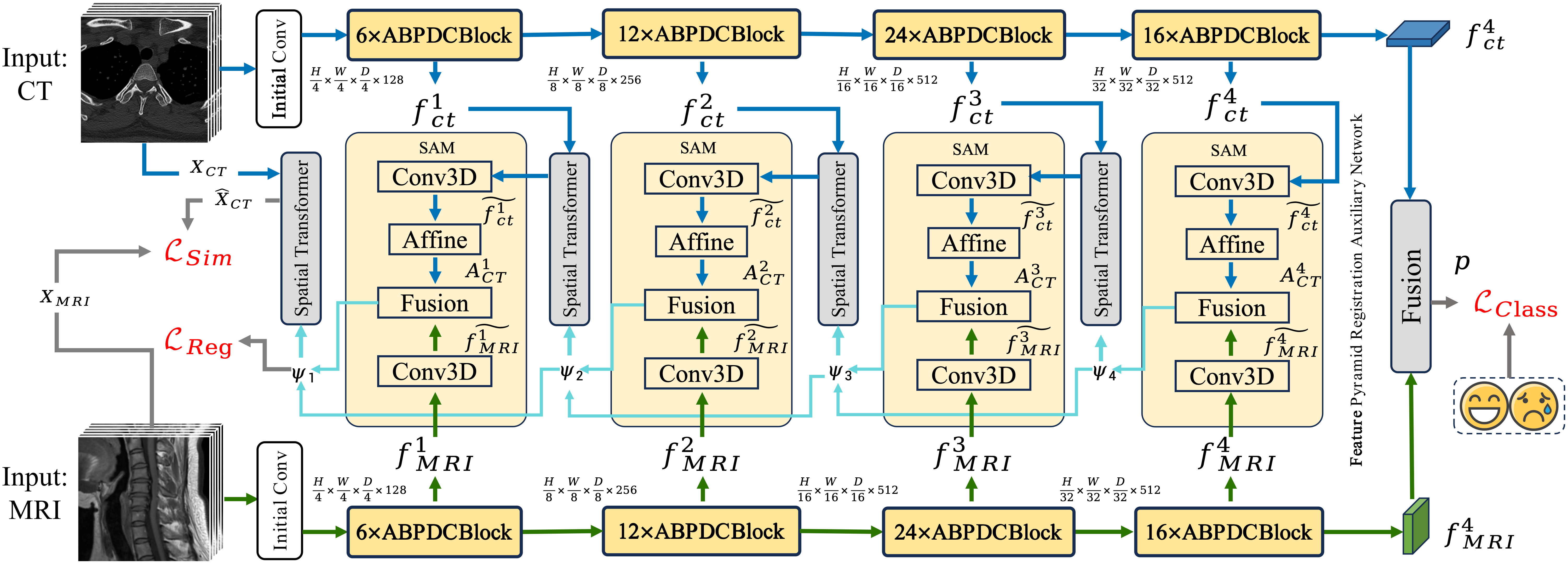}
  \caption{Overall network structure diagram.}
  \label{fig1}
\end{figure*}

\section{Introduction}
\label{sec:intro}
Neck pain is a common symptom of cervical spondylosis and the fourth leading cause of disability worldwide, severely affecting quality of life \cite{spencer2013global}.
Its origin remains debated: some studies implicate soft tissue injuries (e.g., myofascial lesions, intramuscular fat), while others emphasize neural compression by cervical bony structures \cite{qiu2024learning}.
This uncertainty hampers reliable preoperative prediction of postoperative pain relief and complicates surgical decision-making.

Deep learning has shown strong potential in medical imaging, with several studies applying neural networks to spinal surgery planning \cite{ma2023multimodal}.
However, postoperative neck pain prediction remains underexplored: most studies rely solely on clinical data with traditional machine learning \cite{seo2024machine}, neglecting imaging–pain associations.
While image-based models have been studied for other spinal complications, multimodal fusion for pain prediction is rare, and generalizability remains limited.
The recently released MMCSD dataset \cite{yu2025multi} offers paired MRI/CT with pain outcomes, but existing works focus mainly on dataset construction rather than specialized multimodal modelling.

Meanwhile, advances in difference convolution offer a potential solution for multimodal feature integration. 
Originating from Local Binary Patterns (LBP) \cite{Ojala2001}, Yu et al. introduced Central Difference Convolution (CDC) \cite{Yu2020CDCN} to enhance texture perception, followed by improvements in computational efficiency, temporal modeling, edge detection, and fine-grained semantic segmentation \cite{Tan2022SDN}. 
CDC has also been applied to 2D medical imaging \cite{Yu2024BGNet}, yet its extension to 3D thick-slice data is challenged by anisotropic resolution and texture distortion, which often introduce additional noise.

In parallel, self-supervised deformable registration has made progress in bridging multimodal gaps by leveraging intrinsic image features as supervision. 
Hu et al. \cite{Hu2019ContrastiveReg} utilized CT/MR key points but were limited to rigid transformations; Pielawski et al. \cite{Pielawski2020CoMIR} shared cross-modal representations through mutual information but lacked end-to-end optimization; Dey et al. \cite{Dey2022ContraReg} improved accuracy via contrastive learning yet required paired samples. 
Recent cascaded frameworks such as NICE-Net and CorrMLP\cite{Meng2024CorrMLP}demonstrate the effectiveness of coarse-to-fine strategies for brain and cardiac MRI registration. 
Nevertheless, spatial discrepancies between axial CT and sagittal MRI in cervical imaging remain underexplored, further hindering effective multimodal fusion.

To address these gaps, we propose a multimodal deep learning framework that integrates MRI and CT to improve pain-prediction accuracy and provide insights into pain localization. 
This task faces two primary challenges:
(1) Imaging differences: CT and MRI are generated by distinct acquisition principles, resulting in markedly different grayscale distributions and texture patterns. Such modality-specific biases hinder direct feature correspondence and complicate semantic information extraction in multimodal fusion;
(2) Spatial discrepancies: In most clinical protocols, CT is acquired in the axial plane (head-to-foot), whereas MRI is typically obtained in the sagittal plane (left-right). These orthogonal orientations and inconsistent anatomical detail levels make precise voxel-level alignment difficult and limit the effectiveness of cross-modal representation learning.

To address these two problems, we develop two tailored solutions: 
(1) Adaptive Bidirectional Pyramid Difference Convolution (ABPDC), which extends difference convolution to thick-slice 3D images with anisotropic resolution and dynamically adapts coefficients to local gradients for robust multimodal fusion;  
(2) Feature Pyramid Registration Auxiliary Network (FPRAN), a coarse-to-fine alignment strategy that unifies feature spaces across modalities, mitigating scan-orientation inconsistencies.\\
Our contributions are summarized as follows: 
\begin{itemize}[leftmargin=*, itemsep=0pt, parsep=0pt, topsep=0pt]
\item We present the first multimodal neural network that directly associates preoperative MRI/CT images with postoperative neck pain, offering practical support for preoperative clinical decision-making.  
\item We design two complementary modules: ABPDC to alleviate inter-modal imaging differences via adaptive texture fusion and FPRAN to reduce spatial discrepancies through coarse-to-fine registration, together enabling robust multimodal integration.  
\item Our framework achieves state-of-the-art performance on the MMCSD dataset, with prediction accuracy from 76\% to 82\%.  
\end{itemize}

\section{METHODS}
\label{sec:method}
This section presents the proposed method. Section 2.1 introduces the overall framework, Section 2.2 describes the Adaptive Bidirectional Pyramid Difference Convolution (ABPDC), Section 2.3 explains the Feature Pyramid Registration Auxiliary Network (FPRAN), and Section 2.4 defines the loss function.

\subsection{Model Overview}
Data related to neck pain includes axial CT (bone/soft-tissue windows) and sagittal/axial MRI sequences. As shown in Fig.~1, two backbone networks process images from different modalities separately, where standard convolutions are replaced with ABPDC blocks. Given CT and MRI inputs $X_{CT}$ and $X_{MRI}$, the network outputs multi-scale feature pyramids $F_{\mathrm{CT}} = \{ f_{\mathrm{CT}}^{i} \}_{i=1}^{4}$ and $F_{\mathrm{MRI}} = \{ f_{\mathrm{MRI}}^{i} \}_{i=1}^{4}$, where $i$ denotes the pyramid levels from fine to coarse.

FPRAN aligns the feature pyramids from coarse to fine. Starting from $f_{\mathrm{CT}}^{4}$ and $f_{\mathrm{MRI}}^{4}$, the Spatial Alignment Module (SAM) estimates displacement $\psi_4$, which is then propagated to finer levels through a Spatial Transformer Network (STN) \cite{Jaderberg2015STN}. At each level $i$, the displacement is updated by:
\[
\psi_i = 
\begin{cases}
\mathrm{SAM}(f_{\mathrm{CT}}^{i}, f_{\mathrm{MRI}}^{i}), & i=4,\\[0.3em]
\mathrm{SAM}(f_{\mathrm{CT}}^{i}\circ\psi_{i+1}, f_{\mathrm{MRI}}^{i}) + \psi_{i+1}, & i=1,2,3,
\end{cases}
\]
where $\circ$ denotes STN-based warping. 

The final deformation $\psi_1$ warps $X_{CT}$ to $\widehat{X}_{CT}$. Meanwhile, $f_{\mathrm{CT}}^{4}$ and $f_{\mathrm{MRI}}^{4}$ are concatenated and passed to a linear classifier:
\[
\mathbf{p} = \mathbf{W}\!\begin{bmatrix} f^{4}_{\mathrm{CT}} \\ f^{4}_{\mathrm{MRI}} \end{bmatrix} + \mathbf{b},
\]
yielding the prediction vector $\mathbf{p}\in\mathbb{R}^{C}$. Finally, $\mathbf{p}$, $\widehat{X}_{CT}$, and $\psi_1$ are jointly used in the loss function (Section~2.4), enabling the network to simultaneously optimize classification and cross-modal alignment.

\subsection{Adaptive Bidirectional Pyramid Difference Convolution}
Due to anisotropic voxel spacing, medical volumes exhibit richer textures in the x–y plane than along the z-axis. ABPDC therefore emphasizes x–y features by contracting the receptive field pyramidally along $z$ (Fig.~2). For a kernel size $K$, the valid set of pixels is:
\[
S=\{(x,y,z)\mid |z|\le \lfloor K/2\rfloor,\; |x|,|y|\le \lfloor K/2\rfloor-|z|\}.
\]

ABPDC combines standard convolution and difference convolution with adaptive weights $\theta_{adp}$.
Since sharper textures are better captured by standard convolution, while blurred regions require stronger difference enhancement, the weight is adaptively controlled by local texture strength.
To quantify this property, we define the local texture strength $G$ using a 3D Sobel operator, followed by convolution, average pooling, and a sigmoid normalization.
Given an input feature patch $T_{z,x,y}$ centered at $T_c$, ABPDC is formulated as:

\[
\theta_{adp} = 1-\sigma\bigl(\text{Conv}_{1\times1\times1}(\text{Pool}_c(G))\bigr),
\]

\[
\mathrm{ABPDC}(T)=(1-\theta_{adp})\!\!\sum_{C} wT + \theta_{adp}\!\!\sum_{S} w(T-T_{c}),
\]

where $C$ is the full convolutional kernel region, and $S$ is the pyramid-shrunk subset defined by the adaptive difference operation.

\begin{figure}
\centering
\includegraphics[width=1\linewidth]{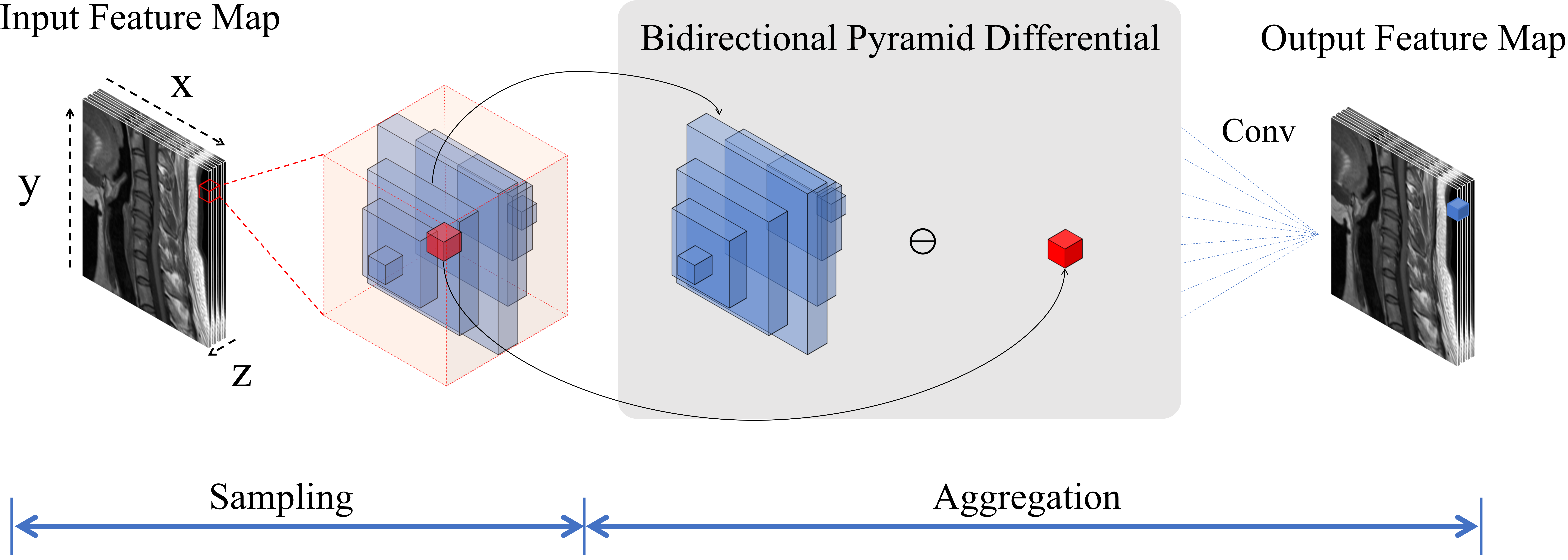}
\caption{Architecture of ABPDC module}
\label{fig2}
\end{figure}

\subsection{Feature Pyramid Registration Auxiliary Network}
FPRAN is designed to progressively align CT and MRI features across multiple scales. It consists of four SAM modules corresponding to the four pyramid levels.  

At the coarsest level, the backbone features $f_{\mathrm{CT}}^{4}$ and $f_{\mathrm{MRI}}^{4}$ are first enhanced by 3D convolution to produce learnable spatial representations $\widetilde f^{4}_{CT}$ and $\widetilde f^{4}_{MRI}$. CT features are then transformed from the axial plane to the sagittal plane by a differentiable affine transformation:
\[
A_{CT}^{4}=\text{Affine}(A,\widetilde f^{4}_{CT}),\quad 
A=R\cdot L(\widetilde f^{4}_{CT}),
\]
where $R$ is the base rotation matrix and $L$ denotes a learnable refinement network of three convolutional layers. This ensures the CT features are geometrically consistent with MRI features.  

The transformed $A_{CT}^{4}$ and $\widetilde f^{4}_{MRI}$ are then fused via a cross-attention mechanism to generate the displacement field $\psi_4$:
\[
\psi_4 = \text{ATT}(A_{CT}^{4}, \widetilde f^{4}_{MRI}).
\]

Next, $f_{\mathrm{CT}}^{3}$ is warped with $\psi_4$ through the STN \cite{Jaderberg2015STN} and fed into the third-layer SAM, which produces $\psi_3=\mathrm{SAM}(f_{\mathrm{CT}}^{3}\!\circ\!\psi_4, f_{\mathrm{MRI}}^{3})+\psi_4$. This refinement process is repeated down to the first level, yielding the final deformation $\psi_1$.

Finally, the warped CT image is obtained as:
\[
\widehat{X}_{CT} = X_{CT}\circ\psi_1,
\]
where $\circ$ denotes STN-based transformation. The similarity between $\widehat{X}_{CT}$ and $X_{MRI}$ is then enforced by the loss function (Section~3.4).

\subsection{Loss Function}
The overall objective combines three terms: classification for prediction accuracy, similarity for cross-modal alignment, and regularization for deformation smoothness:
\[
\mathcal{L}_{\text{All}}=\mathcal{L}_{\text{Class}}+\mathcal{L}_{\text{Sim}}+\mathcal{L}_{\text{Reg}}.
\]
Classification is optimized with cross-entropy,
\[
\mathcal{L}_{\mathrm{Class}}=-\sum_{i=1}^{2} y_i \log(p_i),
\]
similarity is enforced by normalized cross-correlation (NCC),
\[
\mathcal{L}_{\mathrm{Sim}}=-\tfrac{1}{N}\sum_{\Omega}\operatorname{NCC}(X_{MRI},\widehat X_{CT}),
\]
and smoothness is encouraged by penalizing gradients of the displacement field,
\[
\mathcal{L}_{\mathrm{Reg}}=\tfrac{1}{3}\big(\mathbb{E}[\psi_x^2]+\mathbb{E}[\psi_y^2]+\mathbb{E}[\psi_z^2]\big).
\]

\section{EXPERIMENTS}
\label{sec:exp}

\subsection{Dataset and Metrics}

This study is based on the Multi-modal and Multi-view Cervical Spondylosis Imaging Dataset (MMCSD), which contains 250 patients who underwent anterior cervical surgery. Each case includes axial CT bone-window, axial CT soft-tissue-window, axial MRI T2-weighted, and sagittal MRI T1/T2-weighted images. Postoperative Visual Analog Scale (VAS) scores were collected, and patients were divided into pain and non-pain groups. The original train/test split of the dataset was adopted.
Model performance was evaluated using Accuracy, F1-score (harmonic mean of precision and recall), and AUC (area under the ROC curve). 

\subsection{Implementation Details}
The model was implemented using PyTorch and MONAI. All experiments employed the AdamW optimizer with a batch size of 16 and an initial learning rate of $1\times10^{-4}$, decayed by Step Decay (0.9) over 100 epochs. Training was conducted on two NVIDIA RTX 3090 GPUs.

\begin{figure}
\setlength{\abovecaptionskip}{0pt} % 
\setlength{\belowcaptionskip}{0pt} % 
\includegraphics[width=1\linewidth]{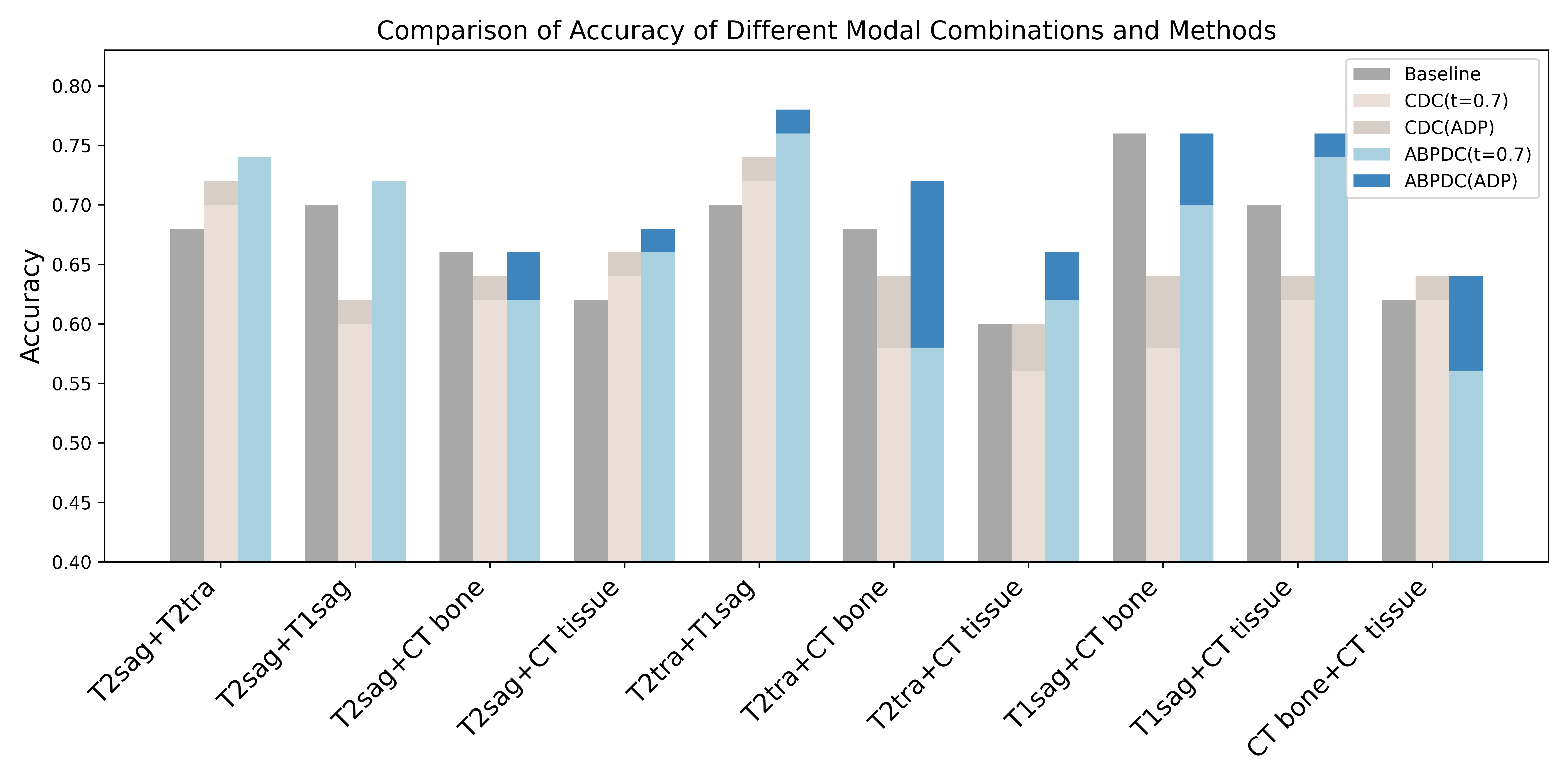}
\caption{Comparison of accuracies for standard convolution, CDC, and ABPDC across modality combinations.}

\label{fig3}
\end{figure}

\subsection{Experiments and Discussions of ABPDC}
We compared standard convolution, CDC, and ABPDC under the CDC hyperparameter $\theta=0.7$ \cite{Yu2020CDCN}. As shown in Fig.~3, both CDC and ABPDC generally outperform standard convolution, with ABPDC consistently superior to CDC. In T2sag+T2tra, T2sag+CT tissue, and T2tra+T1sag, ABPDC achieves the best performance, while in T2sag+T1sag, T2tra+ CT tissue, and T1sag+CT tissue, CDC underperforms standard convolution, but ABPDC still brings improvements. For modality pairs involving CT bone, however, both methods fall below standard convolution.

To mitigate this, we incorporated the adaptive differential weighting algorithm into CDC and ABPDC, which enhanced accuracy in most modality pairs and yielded the largest gains on CT bone. Parameter-tuning further verified that the adaptively selected $\theta$ was close to the optimal value. These results demonstrate the robustness of ABPDC in multimodal fusion.

\begin{table}[htbp]
\captionsetup{skip=10pt} % 
\caption{Results of multi-view ablation experiments with ABPDC (A) and FPRAN (F).}
\centering
\large
\resizebox{\linewidth}{!}{%
\begin{tabular}{l|c|c|c|c|c}
\hline
Sequence & A & F & Accuracy & F1 Score & AUC \\
\hline
\multirow{4}{*}{T2sag+T2tra} & & & 68.00 & 68.00 & 68.11 \\
 & & \checkmark & 70.00(+2.00) & 69.39(+1.39) & 69.23(+1.12) \\
 & \checkmark & & 74.00(+6.00) & 75.50(+7.50) & 76.40(+8.29) \\
 & \checkmark & \checkmark & 76.00(+8.00) & 76.00(+8.00) & 74.83(+6.72) \\
\hline
\multirow{4}{*}{T2sag+CTbone} & & & 66.00 & 69.09 & 65.71 \\
 & & \checkmark & 74.00(+8.00) & 76.36(+7.27) & 78.53(+12.82) \\
 & \checkmark & & 66.00(+0.00) & 63.82(-5.27) & 66.83(+1.12) \\
 & \checkmark & \checkmark & 74.00(+8.00) & 73.46(+4.37) & 74.86(+9.15) \\
\hline
\multirow{4}{*}{T2sag+CTtissue} & & & 62.00 & 59.57 & 61.94 \\
 & & \checkmark & 68.00(+6.00) & 63.64(+4.07) & 63.62(+1.68) \\
 & \checkmark & & 68.00(+6.00) & 66.60(+7.03) & 64.50(+2.56) \\
 & \checkmark & \checkmark & 70.00(+8.00) & 70.58(+11.01) & 72.37(+10.43) \\
\hline
\multirow{4}{*}{T1sag+T2tra} & & & 70.00 & 73.68 & 72.76 \\
 & & \checkmark & 74.00(+4.00) & 73.46(-0.22) & 76.83(+4.07) \\
 & \checkmark & & 78.00(+8.00) & 76.59(+2.91) & 77.90(+5.14) \\
 & \checkmark & \checkmark & 80.00(+10.00) & 80.77(+7.09) & 82.64(+9.88) \\
\hline
\multirow{4}{*}{T1sag+CTbone} & & & 76.00 & 77.78 & 72.76 \\
 & & \checkmark & 78.00(+2.00) & 76.59(-1.19) & 75.63(+3.00) \\
 & \checkmark & & 76.00(+0.00) & 76.00(-1.78) & 77.84(+5.08) \\
 & \checkmark & \checkmark & 82.00(+6.00) & 83.01(+5.23) & 80.35(+7.59) \\
\hline
\multirow{4}{*}{T1sag+CTtissue} & & & 70.00 & 68.09 & 71.96 \\
 & & \checkmark & 74.00(+4.00) & 76.36(+8.27) & 77.35(+5.39) \\
 & \checkmark & & 76.00(+6.00) & 76.92(+8.83) & 74.90(+2.94) \\
 & \checkmark & \checkmark & 78.00(+8.00) & 78.43(+10.34) & 80.24(+8.28) \\
\hline
\end{tabular}
}
\end{table}

\subsection{Experiments and Discussions of FPRAN}
To evaluate FPRAN for multi-view fusion, we paired sagittal MRI (T1/T2) with axial MRI-T2, CT bone, and CT soft tissue, yielding six modality combinations. As shown in Table~1, FPRAN consistently achieves superior performance.

We further conducted ablation studies to examine the composability of ABPDC and FPRAN. Results show that integrating the two does not compromise performance; instead, they work synergistically. Five-fold cross-validation was used to evaluate generalization.

Overall, ABPDC and FPRAN achieve state-of-the-art results on this dataset, with our method reaching 82\% prediction accuracy.

\subsection{Comparison with Backbone Networks}
We further compared our method with commonly used generic backbones (e.g., ResNet, ViT, EfficientNet, DenseNet) and representative medical imaging backbones (e.g., PBTC TransNet, CoPAS). As shown in Table~2, generic backbones exhibit limited performance, while medical backbones provide stronger results. Nevertheless, our method consistently outperforms all existing approaches across accuracy, AUC, and F1-score, demonstrating clear superiority and strong adaptability to the neck pain dataset.

\begin{table}[h!]
\caption{Comparative experiments on backbone networks}
\centering
\large
\resizebox{\linewidth}{!}{%
\begin{tabular}{l|ccc|ccc}
\toprule
\multirow{2}{*}{method} & \multicolumn{3}{c|}{T1sag+T2tra} & \multicolumn{3}{c}{T1sag+CTbone} \\
\cmidrule(lr){2-4} \cmidrule(lr){5-7}
 & ACC & AUC & F1 Score & ACC & AUC & F1 Score \\
\midrule
Resnet\cite{He_2016_CVPR}   &  60.0   &  66.3   &  52.3   &  64.0   &   66.1  &   57.1  \\
Vit\cite{dosovitskiy2020image}      &  58.0   &   55.1  &   67.6  &  58.0   &  56.7   &   48.7  \\
Efficientnet\cite{tan2019rethinking} &  64.0   &   56.0  &   71.87  &  64.0   &  68.2   &   71.8  \\
Densenet\cite{huang2017densely} &  74.0   &  74.2   &  73.5   &  66.0   &   66.2  &   65.3  \\
\cmidrule(lr){1-1}\cmidrule(lr){2-4} \cmidrule(lr){5-7}
PBTC TransNet\cite{song2024deep} &  76.0   &  79.5   &  76.0  &   70.0  &   \textbf{86.5}  &  76.9   \\
CoPAS\cite{qiu2024learning}     &  74.0  &  81.7   &  72.3   &  74.0   &   82.3  &   72.3  \\
\textbf{Ours} & \textbf{80.0} & \textbf{82.6} & \textbf{80.7} & \textbf{82.0} & 80.3 & \textbf{83.0} \\

\bottomrule
\end{tabular}
}
\end{table}

\section{Conclusion}
This study, based on the MMCSD dataset, proposes a neural network that integrates multi-modal and cross-view medical images to predict postoperative neck pain in cervical spondylosis. The ABPDC module addresses modality differences in image fusion, while the FPRAN module corrects spatial misalignment from inconsistent scanning orientations. Experimental results show that the proposed method outperforms baseline models \cite{yu2025multi} across multiple modality combinations, achieving up to 82\% prediction accuracy. Ablation studies further confirm the independent contributions of each sub-module. These findings highlight the potential of multimodal deep learning to support preoperative decision-making in cervical spondylosis.\\

\bibliographystyle{IEEEbib}
\bibliography{refs}

\begin{thebibliography}{10}

\bibitem{spencer2013global}
Stuart Spencer,
\newblock ``Global burden of disease 2010 study: a personal reflection,''
\newblock {\em Global Cardiology Science and Practice}, vol. 2013, no. 2, pp.
  15, 2013.

\bibitem{qiu2024learning}
Zelin Qiu, Zhuoyao Xie, Huangjing Lin, Yanwen Li, Qiang Ye, Menghong Wang,
  Shisi Li, Yinghua Zhao, and Hao Chen,
\newblock ``Learning co-plane attention across mri sequences for diagnosing
  twelve types of knee abnormalities,''
\newblock {\em Nature Communications}, vol. 15, no. 1, pp. 7637, 2024.

\bibitem{ma2023multimodal}
Chao Ma, Liyang Wang, Dengpan Song, Chuntian Gao, Linkai Jing, Yang Lu,
  Dongkang Liu, Weitao Man, Kaiyuan Yang, Zhe Meng, et~al.,
\newblock ``Multimodal-based machine learning strategy for accurate and
  non-invasive prediction of intramedullary glioma grade and mutation status of
  molecular markers: a retrospective study,''
\newblock {\em BMC medicine}, vol. 21, no. 1, pp. 198, 2023.

\bibitem{seo2024machine}
Yechan Seo, Seoi Jeong, Siyoung Lee, Tae-Shin Kim, Jun-Hoe Kim, Chun~Kee Chung,
  Chang-Hyun Lee, John~M Rhee, Hyoun-Joong Kong, and Chi~Heon Kim,
\newblock ``Machine-learning-based models for the optimization of post-cervical
  spinal laminoplasty outpatient follow-up schedules,''
\newblock {\em BMC Medical Informatics and Decision Making}, vol. 24, no. 1,
  pp. 278, 2024.

\bibitem{yu2025multi}
Qi-Shuai Yu, Jing-Yang Shan, Jie Ma, Gan Gao, Ben-Zhang Tao, Guang-Yu Qiao,
  Jian-Ning Zhang, Ting Wang, Yong-Fei Zhao, Xiao-Lin Qin, et~al.,
\newblock ``Multi-modal and multi-view cervical spondylosis imaging dataset,''
\newblock {\em Scientific Data}, vol. 12, no. 1, pp. 1080, 2025.

\bibitem{Ojala2001}
T.~Ojala, K.~Valkealahti, E.~Oja, and M.~Pietik{\"a}inen,
\newblock ``Texture discrimination with multi-dimensional distributions of
  signed gray level differences,''
\newblock {\em Pattern Recognition}, vol. 34, pp. 727--739, 2001.

\bibitem{Yu2020CDCN}
Z.~Yu, C.~Zhao, Z.~Wang, et~al.,
\newblock ``Searching central difference convolutional networks for face
  anti-spoofing,''
\newblock in {\em Proceedings of the IEEE/CVF Conference on Computer Vision and
  Pattern Recognition (CVPR)}, 2020, pp. 5295--5305.

\bibitem{Tan2022SDN}
H.~Tan, S.~Wu, and J.~Pi,
\newblock ``Semantic diffusion network for semantic segmentation,''
\newblock in {\em Advances in Neural Information Processing Systems 35
  (NeurIPS)}, 2022, pp. 8702--8716.

\bibitem{Yu2024BGNet}
L.~Yu, W.~Min, and S.~Wang,
\newblock ``Boundary-aware gradient operator network for medical image
  segmentation,''
\newblock {\em IEEE Journal of Biomedical and Health Informatics}, vol. 28, no.
  8, pp. 4711--4723, 2024.

\bibitem{Hu2019ContrastiveReg}
J.~Hu, S.~Sun, X.~Yang, et~al.,
\newblock ``Towards accurate and robust multi-modal medical image registration
  using contrastive metric learning,''
\newblock {\em IEEE Access}, vol. 7, pp. 132816--132827, 2019.

\bibitem{Pielawski2020CoMIR}
N.~Pielawski, E.~Wetzer, J.~{\"O}fverstedt, et~al.,
\newblock ``Comir: Contrastive multimodal image representation for
  registration,''
\newblock in {\em Advances in Neural Information Processing Systems 33
  (NeurIPS)}, 2020,
\newblock Paper~id d6428eecbe0f7dff83fc607c5044b2b9.

\bibitem{Dey2022ContraReg}
N.~Dey, J.~Schlemper, S.~S.~M. Salehi, et~al.,
\newblock ``Contrareg: Contrastive learning of multi-modality unsupervised
  deformable image registration,''
\newblock in {\em Medical Image Computing and Computer-Assisted Intervention
  (MICCAI 2022)}, 2022, pp. 66--77.

\bibitem{Meng2024CorrMLP}
M.~Meng, D.~Feng, L.~Bi, and J.~Kim,
\newblock ``Correlation-aware coarse-to-fine mlps for deformable medical image
  registration,''
\newblock in {\em Proceedings of the IEEE/CVF Conference on Computer Vision and
  Pattern Recognition (CVPR)}, 2024, pp. 9645--9654.

\bibitem{Jaderberg2015STN}
M.~Jaderberg, K.~Simonyan, and A.~Zisserman,
\newblock ``Spatial transformer networks,''
\newblock in {\em Advances in Neural Information Processing Systems 28 (NIPS)},
  2015, pp. 2017--2025.

\bibitem{He_2016_CVPR}
Kaiming He, Xiangyu Zhang, Shaoqing Ren, and Jian Sun,
\newblock ``Deep residual learning for image recognition,''
\newblock in {\em Proceedings of the IEEE Conference on Computer Vision and
  Pattern Recognition (CVPR)}, June 2016.

\bibitem{dosovitskiy2020image}
Alexey Dosovitskiy, Lucas Beyer, Alexander Kolesnikov, Dirk Weissenborn,
  Xiaohua Zhai, Thomas Unterthiner, Mostafa Dehghani, Matthias Minderer, Georg
  Heigold, Sylvain Gelly, et~al.,
\newblock ``An image is worth 16x16 words: Transformers for image recognition
  at scale,''
\newblock {\em arXiv preprint arXiv:2010.11929}, 2020.

\bibitem{tan2019rethinking}
Mingxing Tan, Q~Efficientnet Le, et~al.,
\newblock ``Rethinking model scaling for convolutional neural networks,''
\newblock in {\em Proceedings of the International conference on machine
  learning, Long Beach, CA, USA}, 2019, vol.~15.

\bibitem{huang2017densely}
Gao Huang, Zhuang Liu, Laurens Van Der~Maaten, and Kilian~Q Weinberger,
\newblock ``Densely connected convolutional networks,''
\newblock in {\em Proceedings of the IEEE conference on computer vision and
  pattern recognition}, 2017, pp. 4700--4708.

\bibitem{song2024deep}
Liwen Song, Chuanpu Li, Lilian Tan, Menghong Wang, Xiaqing Chen, Qiang Ye,
  Shisi Li, Rui Zhang, Qinghai Zeng, Zhuoyao Xie, et~al.,
\newblock ``A deep learning model to enhance the classification of primary bone
  tumors based on incomplete multimodal images in x-ray, ct, and mri,''
\newblock {\em Cancer Imaging}, vol. 24, no. 1, pp. 135, 2024.

\end{thebibliography}

\end{document}